\documentclass[conference]{IEEEtran}
\usepackage{graphicx}
\usepackage{cite}
\usepackage{amsmath}
\usepackage{enumitem}  

\usepackage{algorithm}
\usepackage{algpseudocode}
\usepackage{url}
\usepackage{booktabs}
\usepackage{multirow}
\usepackage{rotating}
\usepackage{array}
\usepackage{xcolor}
\usepackage{comment}
\usepackage{listings}

\begin{document}

\title{Augmented Fine-Tuned LLMs for Enhanced Recruitment Automation}

\author{
\IEEEauthorblockN{{Mohamed T. Younes\large\IEEEauthorrefmark{1},
  Omar Walid\IEEEauthorrefmark{1},
Khaled Shaban\IEEEauthorrefmark{2},
Ali Hamdi\IEEEauthorrefmark{1},
Mai Hassan\IEEEauthorrefmark{1}}}

\IEEEauthorblockA{\IEEEauthorrefmark{1}\large\textit{Dept. of Computer Science, MSA University, Giza, Egypt} \\
\{mohamed.tarek61  , omar.walid2, ahamdi, maisalem\}@msa.edu.eg}

\IEEEauthorblockA{\IEEEauthorrefmark{2}\large\textit{Dept. of Computer Science, Qatar University, Doha, Qatar} \\
khaled.shaban@qu.edu.qa}
}

\maketitle
\begin{abstract}
This paper presents a novel approach to recruitment automation. Large Language Models (LLMs) were fine-tuned to improve accuracy and efficiency. Building upon our previous work on the Multilayer Large Language Model-Based Robotic Process Automation Applicant Tracking (MLAR) system \cite{younes2025mlar} . This work introduces a novel methodology. Training fine-tuned LLMs specifically tuned for recruitment tasks. The proposed framework addresses the limitations of generic LLMs by creating a synthetic dataset that uses a standardized JSON format. This helps ensure consistency and scalability. In addition to the synthetic data set, the resumes were parsed using DeepSeek, a high-parameter LLM. The resumes were parsed into the same structured JSON format and placed in the training set. This will help improve data diversity and realism. Through experimentation, we demonstrate significant improvements in performance metrics, such as exact match, F1 score, BLEU score, ROUGE score, and overall similarity compared to base models and other state-of-the-art LLMs. In particular, the fine-tuned Phi-4 model achieved the highest F1 score of 90.62\%, indicating exceptional precision and recall in recruitment tasks. This study highlights the potential of fine-tuned LLMs. Furthermore, it will revolutionize recruitment workflows by providing more accurate candidate-job matching.
\end{abstract}

\textbf{Index Terms}---Applicant Tracking System (ATS), Robotic Process Automation (RPA), Large Language Models (LLMs), Fine-Tuning.

\section{Introduction}
The recruitment process is a key company task. This task faces growing challenges. It needs quick review of many applications. The process must keep high selection rules. Today’s hiring world needs better systems. These systems handle hundreds of candidates per job. They provide fair checks of applicant skills. They provide correct checks of applicant skills. Traditional Applicant Tracking Systems (ATS), despite their implementation in most organizations, continue to face limitations in meeting these demands. Their reliance on simplistic keyword matching and manual screening approaches often introduces biases, inconsistencies, and time consumptions, ultimately leading to poor hiring decisions.

New advancements in artificial intelligence has changed recruitment workflows a lot. Tools like robotic process automation (RPA) and large language models (LLMs) lead this change. These tools can automate and improve talent acquisition steps. They handle everything from resume screening to full candidate reviews. But many current systems use generic language models (non finetuned LLMS). These models don’t focus on recruitment needs. They often struggle to understand the tricky language in resumes and job descriptions. Resulting in inaccurate information extraction and poor candidate-position matching. The structural complexity and variability of resume formats further worsen these challenges, as most systems struggle to consistently parse relevant information from unstructured documents.

Our prior research addressed some of these limitations through the development of MLAR (Multilayer Large Language Model-Based Robotic Process Automation for Applicant Tracking) , a system that demonstrated the potential of combining RPA with LLMs for recruitment automation \cite{younes2025mlar}. MLAR showed measurable improvements over conventional ATS platforms, particularly in handling diverse resume formats and reducing processing times. However, its dependence on base language models without domain-specific adaptation revealed ongoing issues in correctly understanding technical skills and job experiences tied to hiring needs.

This paper presents an advance over our previous work by introducing a novel methodology to fine-tune LLMs specifically for recruitment tasks. Our approach combines several key contributions that address current limitations in the field:
\begin{enumerate}
    \item Develop a comprehensive training corpus that merges parsed real resume data with generated synthetic examples, all standardized through a strict JSON schema. This hybrid data set ensures both the realism of actual recruitment data and the coverage of edge cases through synthetic generation.
    \item Implement specialized fine-tuning of multiple state-of-the-art open-weight LLMs, including LLaMA3.1 8B, Mistral 7B, Phi-4 14B, and Gemma2 9B, using parameter-efficient adaptation techniques.
    \item Establish a strong testing system that checks model performance using both standard language metrics and hiring-specific measures.
\end{enumerate}
Combined synthetically generated resumes and real-world examples parsed with DeepSeek in a strict JSON format. Fine-tuning of LLaMA3.1 8B, Mistral 7B, Phi-4 14B and Gemma2 9B for recruitment-specific tasks. Comprehensive evaluation using Exact Match (EM), F1 score, BLEU score, ROUGE score and overall similarity. Integration with the MLAR framework to assess real-world impact. \cite{younes2025mlar} Algorithmic formalization through pseudocode for data set generation, model fine-tuning, and evaluation.

This research impacts several aspects of modern recruitment systems. We developed domain-specific language models for recruitment tasks. These models improve the parsing accuracy of candidate qualifications. They also enhance the precision of candidate-job matching. Our solution addresses critical data privacy issues in human resources systems. The standardized JSON output format enables smooth integration with existing HR software frameworks. Additionally, our structured evaluation methodology offers key insights. These insights benefit researchers and professionals implementing AI-driven recruitment solutions.

The remainder of this paper is organized as follows: Section~\ref{relatedwork} reviews related work on robotic process automation, large language models, and fine-tuning techniques for recruitment automation. Section~\ref{methodology} outlines the problem formulation, dataset creation, model fine-tuning process, and evaluation metrics used in this study. Section~\ref{experimentaldesign} describes the experimental design, including the system workflow and benchmarking setup. Section~\ref{results} presents the performance comparison of fine-tuned models against base models, along with key observations. Finally, Section~\ref{conclusion} concludes the paper with insights on the effectiveness of the proposed approach and potential future research directions.

\begin{figure}[t]    
\centering\includegraphics[scale=0.5]{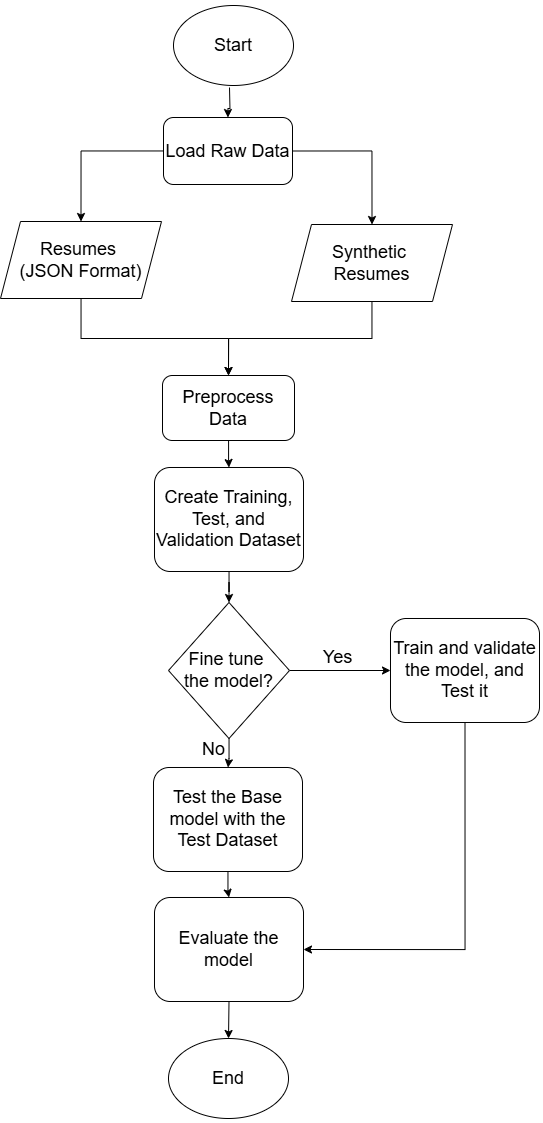}
    \caption{Flow Diagram, outlining the key steps from data preparation to model evaluation.}
    \label{fig:Sysflowchart}
\end{figure}

\section{Related Work} \label{relatedwork}
\subsection{Robotic Process Automation in Recruitment}
Robotic Process Automation (RPA) has emerged as a powerful tool for automating repetitive tasks in recruitment. These tasks include resume parsing, email management, and workflow coordination. Popular RPA platforms like UiPath and Automation Anywhere provide robust tools for task automation in recruitment. These platforms offer user-friendly interfaces and analytics capabilities but often require significant customization for recruitment-specific tasks. For instance, research by Balasundaram and Venkatagiri \cite{Balasundaram2020} illustrates how structured RPA implementations can streamline HR processes, including recruitment, by automating repetitive workflows. Similarly, Laumer et al. \cite{Laumer2015} emphasize the impact of integrating business process management with ATS to enhance recruitment efficiency, highlighting the potential of RPA in this domain.

Research has also shown that integrating RPA with AI technologies can significantly enhance recruitment workflows. Studies have demonstrated how RPA combined with AI tools could automate resume parsing and categorization. This automation applies methods like Naive Bayes classifiers and Named Entity Recognition (NER) \cite{Nawaz2019}. Our prior work introduced the MLAR system , which merges the RPA architecture with LLMs. \cite{younes2025mlar} It improved resume parsing and reduced processing times. These results show the value of such integrations. However, these advancements often lack deeper contextual understanding, underscoring the need for smarter decision-making abilities in automated systems, which our current work addresses through fine-tuning.

\begin{figure*}[t]    
\centering\includegraphics[scale=0.15]{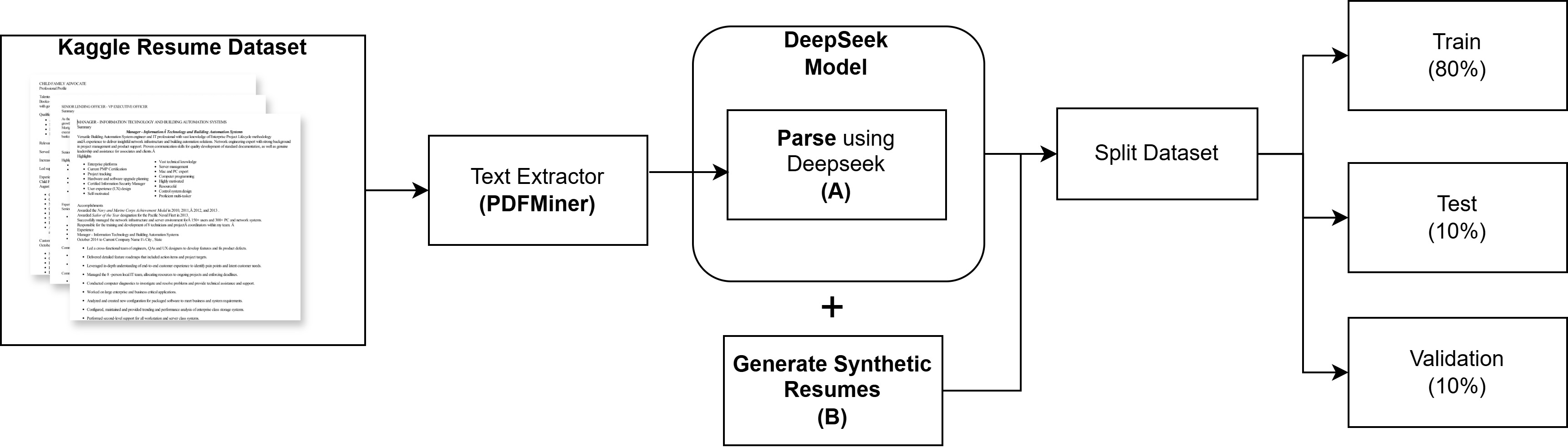}
    \caption{Flow Diagram, outlining the key steps in the creation of the hybrid dataset}
    \label{fig:datasetcreation}
\end{figure*}

\subsection{Large Language Models in Recruitment}
Recent advancements in Large Language Models (LLMs) have transformed text comprehension and analysis. These advancements have helped enabling more accurate interpretation of unstructured data like resumes and job descriptions. Studies introduced BERT-based frameworks and Gemini LLMs to evaluate resumes and job descriptions, calculating similarity scores to predict candidate suitability. For example, Abdollahnejad et al. \cite{Abdollahnejad2021} proposed a deep learning BERT-based approach that significantly improves person-job fit assessments, outperforming traditional methods. Wang et al. \cite{Wang2023} further provide a comprehensive survey on HR process automation, noting how LLMs enhance candidate-job matching by capturing semantic relationships beyond simple keyword searches.

However, base LLMs face several limitations when applied to recruitment. They often lack the domain-specific understanding required to understand industry terminology, role-specific skills, and the complex structure of resumes. Moreover, off-the-shelf LLMs are prone to generating hallucinated outputs.They also may struggle with consistency when parsing large batches of varied resume formats \cite{Wang2023}. These limitations create a performance bottleneck in real-world recruitment applications. They require the fine-tuning of models on domain-relevant data to ensure reliability, scalability, and fairness, an approach we extend in this work, building on our earlier MLAR framework.

\subsection{Fine-Tuning LLMs for Recruitment}
Fine-tuning LLMs on domain-specific datasets has become a popular approach to improving model performance in targeted applications. Research demonstrated that fine-tuning LLMs on recruitment datasets could significantly enhance their ability to summarize resumes and grade candidates effectively. Vishaline et al. \cite{anmlbased2024} proposed an ML-based resume screening and ranking system that leverages classifiers such as SVM, Naive Bayes, and XGBoost to evaluate candidate suitability. Their study highlights how machine learning models, when integrated into recruitment workflows, can provide structured performance metrics and improve the accuracy of applicant ranking, thereby advancing the effectiveness of automated hiring systems. Similarly, Vagale et al. \cite{Vagale2024} developed ProspectCV, an LLM-based platform that advances CV-job description evaluation, demonstrating the value of fine-tuning for precise matching.

Despite these advancements, there remains a need for more robust and scalable solutions. These solutions should be able to handle the complexities of recruitment workflows. Schaudt and Schlegel \cite{Schaudt2023} argue that combining RPA with fine-tuned AI models addresses many limitations of generic systems, offering a path toward more effective automation. Our previous work on the MLAR system laid the groundwork for such integration, showing initial promise in automating recruitment tasks \cite{younes2025mlar}.  However, it revealed the need for domain-specific fine-tuning. This fine-tuning addresses the shortcomings of generic LLMs. Our current paper fills this gap. It introduces a novel fine-tuning methodology.

\section{Research Methodology} \label{methodology}
\subsection{Problem Formulation}
\label{sec:problem_formulation}

Recruitment automation aims to streamline the hiring process by leveraging advanced natural language processing (NLP) techniques to parse resumes and match candidates to job requirements. However, this task presents several significant challenges that must be addressed to achieve reliable and efficient automation. In this work, we focus on the task of resume parsing, defined as the extraction of structured information from unstructured resume text into a predefined JSON format. Specifically, given an unstructured resume, the goal is to extract key entities such as the candidate's name, contact details, skills, work experience, and education, while ensuring accuracy and consistency across diverse resume formats.

One primary challenge in recruitment automation is the \textit{lack of domain-specific understanding} in generic large language models (LLMs). Models such as BERT~\cite{devlin2019bert} and DeBERTa~\cite{he2021deberta} are typically trained on general-purpose corpora, which limits their ability to comprehend industry-specific terminology and context relevant to recruitment~\cite{gururangan2020don}. For instance, these models may fail to accurately identify and extract nuanced skills (e.g., distinguishing between ``Python programming'' and ``Python scripting'' in a software engineering context) or misinterpret domain-specific jargon, leading to incomplete or erroneous parsing.

Another critical challenge is the \textit{complexity and variability in resume structures}. Resumes exhibit significant diversity in formatting, language, and organization, ranging from highly structured documents with clear headings to unstructured free-text formats with unconventional layouts. For example, a resume may present work experience in a tabular format, as a narrative paragraph, or even in multiple languages, making it difficult for generic LLMs to consistently parse and extract relevant information. This variability often results in missed entities or incorrect mappings, undermining the reliability of automated recruitment systems~\cite{xu2020layoutlm}.

Additionally, generic LLMs frequently suffer from \textit{hallucinated outputs and inconsistency} when parsing diverse resume formats. Hallucination occurs when a model generates fabricated or incorrect information, such as inventing job titles or skills that do not exist in the resume. For example, a generic LLM might hallucinate a skill like ``Java expertise'' for a candidate who only mentioned ``JavaScript'' due to poor contextual understanding. Furthermore, inconsistencies arise when the same model produces varying outputs for similar inputs, such as extracting a candidate’s name correctly in one resume but failing to do so in another with a similar structure. These issues highlight the need for domain-specific fine-tuning to improve the robustness and reliability of LLMs in recruitment tasks~\cite{xiao2021hallucination}.

To address these challenges, our approach involves fine-tuning LLMs on a hybrid dataset tailored to recruitment automation, enabling the models to better understand domain-specific terminology, handle structural variability, and reduce hallucination. We evaluate the effectiveness of our fine-tuning strategy using metrics such as Exact Match and F1 Score, which are standard for assessing information extraction tasks~\cite{wang2020superglue}, as detailed in Section~\ref{sec:Evaluation metrics}.

\subsection{Hybrid Dataset Creation}

\label{sec:dataset_creation}

The dataset used in this research was created through a combination of real-world and synthetic data, as illustrated in (Figure~\ref{fig:datasetcreation}). The process involves extracting text from real-world resumes, parsing the text using DeepSeek, generating synthetic resumes using the DeepSeek model, combining them into a hybrid dataset, and finally splitting the dataset into training, validation, and testing sets.

The real-world dataset is sourced from the Kaggle Resume Dataset~\cite{kaggle_resume_dataset}, which consists of 2,400 resumes in PDF format spanning 24 professions, including Human Resources, Information Technology, Public Relations, and Healthcare. Since the resumes are in PDF format, the first step is to extract the text using PDFMiner, a tool designed for extracting text from PDF files. This step converts the unstructured PDF content into plain text, making it suitable for further processing.

The extracted text is then parsed using DeepSeek~\cite{deepseek_model}, a large language model with 236 billion parameters known for its advanced natural language processing capabilities. DeepSeek was selected for this task due to its ability to handle complex, unstructured text and accurately extract key entities such as name, email, phone, skills, experience, education, and department. Its large parameter size and strong performance in information extraction tasks make it well-suited for generating high-quality reference outputs . The parsed data is structured into a standardized JSON format.

In addition to the real-world resumes, synthetic resumes are generated using the DeepSeek model to augment the dataset. The generation process employs techniques similar to those described by Ratner et al.~\cite{Ratner2017}, who advocate for data programming to efficiently create large training sets in NLP tasks. This step involves creating artificial resumes that adhere to the same JSON structure as the parsed real-world resumes.

The parsed real-world resumes and the synthetic resumes are then combined to form a hybrid dataset. This hybrid approach ensures that the dataset includes both realistic recruitment data and a variety of scenarios, enhancing the diversity and robustness of the training data.

Before splitting, the hybrid dataset was normalized to resolve inconsistencies, such as standardizing date formats and using placeholders like "John Doe" for missing names. The normalized dataset was then split into three subsets: 80\% for training, 10\% for validation, and 10\% for testing.

It is important to note that DeepSeek was used exclusively for parsing the real-world resumes and generating synthetic resumes to establish high-quality reference data. DeepSeek was not included in the subsequent fine-tuned model comparison, which focused on other open-weight models such as LLAMA3.1 8B~\cite{Touvron2023}, Mistral 7B~\cite{Jiang2023}, Phi-4 14B~\cite{phi_model}, and Gemma2 9B~\cite{gemma_model}. These models were selected for their accessibility and architectural diversity, allowing for a broader evaluation of fine-tuning strategies in recruitment automation. By using DeepSeek solely for dataset creation, we ensured a reliable and consistent standard for evaluating the performance of the fine-tuned models.

\begin{algorithm}[t]
\caption{Dataset Creation and Preprocessing}
\begin{enumerate}
    \item \textbf{Step 1: Parse Real Resumes}
    \begin{enumerate}
        \item \textit{Loop through each resume $r_i \in R$.}
        \item \textit{Extract structured fields using DeepSeek.}
        \item \textit{Convert the result to standardized JSON format $j_i$.}
        \item \textit{Add $j_i$ to the real dataset $D_{\text{real}}$.}
    \end{enumerate}
    \item \textbf{Step 2: Generate Synthetic Resumes}
    \begin{enumerate}
        \item \textit{Repeat for $i = 1$ to $m$.}
        \item \textit{Generate a synthetic resume $s_i \gets$ Generate\_resume().}
        \item \textit{Convert $s_i$ to standardized JSON format.}
        \item \textit{Add $s_i$ to the synthetic dataset $D_{\text{synth}}$.}
    \end{enumerate}

    \item \textbf{Step 3: Combine and Normalize}
    \begin{enumerate}
        \item \textit{Merge both datasets: $D_{\text{combined}} \gets D_{\text{real}} \cup D_{\text{synth}}$.}
        \item \textit{Normalize all fields across $D_{\text{combined}}$.}
        \item \textit{Standardize date formats.}
        \item \textit{Unify skill terminology.}
        \item \textit{Address any missing values consistently.}
    \end{enumerate}

    \item \textbf{Step 4: Split Dataset}
    \begin{enumerate}
        \item \textit{Partition $D_{\text{combined}}$ into training, validation, and test sets using an 80/10/10 split.}
        \item \textit{Return the final dataset: $D_{\text{final}} = \{D_{\text{train}}, D_{\text{val}}, D_{\text{test}}\}$.}
    \end{enumerate}
\end{enumerate}
\end{algorithm}

\subsection{Model Fine-Tuning}

For the fine-tuning phase, we selected four advanced LLMs: LLaMA 3.1 8B \cite{Touvron2023}, Mistral 7B \cite{Jiang2023}, Phi-4 14B \cite{phi_model}, and Gemma 2 9B \cite{gemma_model}, chosen for their suitability for proficiency in natural language tasks. These models were fine-tuned on the combined dataset using supervised learning to parse resume texts and extract key information into the standardized JSON format. The fine-tuning process, detailed in Algorithm 2, employed Low-Rank Adaptation (LoRA) to efficiently adapt the pre-trained models, updating a small subset of parameters to reduce computational overhead while achieving significant performance gains.

\begin{algorithm}[htpb]
\caption{Model Fine-Tuning Process}
\begin{enumerate}
    \item \textbf{Initialize Model}
    \begin{enumerate}
        \item \textit{Load model $M$ with pre-trained weights.}
    \end{enumerate}

    \item \textbf{Configure LoRA Parameters}
    \begin{enumerate}
        \item \textit{Set rank $r \gets 16$.}
        \item \textit{Set scaling factor $\alpha \gets 16$.}
        \item \textit{Define target modules: ["q\_proj", "k\_proj", "v\_proj", "o\_proj"].}
    \end{enumerate}

    \item \textbf{Define Training Hyperparameters}
    \begin{enumerate}
        \item \textit{Batch size $\gets 8$.}
        \item \textit{Learning rate $\gets 5 \times 10^{-5}$.}
        \item \textit{Max training steps $\gets 200$.}
        \item \textit{Warmup steps $\gets 5$.}
    \end{enumerate}

    \item \textbf{Training Loop}
    \begin{enumerate}
        \item \textit{For each training step:}
        \begin{enumerate}
            \item \textit{Sample batch $B \gets D_{\text{train}}$.}
            \item \textit{Perform forward pass and compute loss $L$.}
            \item \textit{Run backward pass and update LoRA parameters.}
            \item \textit{If at validation interval:}
            \begin{enumerate}
                \item \textit{Evaluate on $D_{\text{val}}$.}
                \item \textit{Adjust learning rate if necessary.}
            \end{enumerate}
        \end{enumerate}
    \end{enumerate}

    \item \textbf{Return Fine-Tuned Model}
    \begin{enumerate}
        \item \textit{Output $M_{\text{fine-tuned}}$.}
    \end{enumerate}
\end{enumerate}
\end{algorithm}

\subsection{Evaluation Metrics} 
\label{sec:Evaluation metrics}

The performance of the fine-tuned models was evaluated using a suite of metrics tailored to recruitment tasks: Exact Match (EM) using Levenshtein distance ratio; F1 Score with Sentence-BERT embeddings \cite{Reimers2019} for semantic similarity; BLEU Score (smoothed BLEU-4) \cite{Papineni2002} for n-gram overlap; ROUGE Score combining ROUGE-1, ROUGE-2, and ROUGE-L with stemming \cite{Lin2004}; and Overall Similarity as a composite score. The evaluation process, implemented in Python and detailed in Algorithm 3, handles edge cases like missing fields and formatting variations.

\subsubsection{\textbf{Metric Definitions}}
\begin{enumerate}
    \item \textbf{Exact Match (EM):} We compute an approximate exact match score using Levenshtein distance ratio between corresponding fields in the reference and predicted outputs:
    \[
    EM = \frac{1}{|F|} \sum_{f \in F} \text{LevenshteinRatio}(R_f, P_f)
    \]
    where $F$ is the set of all fields, $R_f$ is the reference value for field $f$, and $P_f$ is the predicted value.

    \item \textbf{F1 Score with Semantic Similarity:} We calculate a modified F1 score using SBERT embeddings to capture semantic similarity:
    \[
    \mathrm{F1_{sem}} = \cos(\operatorname{SBERT}(R), \operatorname{SBERT}(P))
    \]
    where $R$ and $P$ are the flattened reference and predicted outputs, respectively.

    \item \textbf{BLEU Score:} We employ the smoothed BLEU-4 metric to evaluate fluency and n-gram overlap:
    \[
    \text{BLEU} = \text{sentence\_bleu}(R, P, \text{smoothing\_function} = \text{method4})
    \]

    \item \textbf{ROUGE Score:} We compute a combined ROUGE score incorporating ROUGE-1, ROUGE-2, and ROUGE-L with stemming:
    \[
    \text{ROUGE} = \frac{1}{3} (\text{ROUGE-1}_f + \text{ROUGE-2}_f + \text{ROUGE-L}_f)
    \]
\end{enumerate}

\begin{table*}[t]\label{Table1}
\centering
\caption{Performance Comparison of Fine-Tuned and Base Models on Recruitment Tasks}
\normalsize
\renewcommand{\arraystretch}{1.2}
\begin{tabular}{|c|c|c|c|c|c|c|}
\hline
& \textbf{Model} & \textbf{Parameters} & \textbf{EM } & \textbf{F1} & \textbf{BLEU} & \textbf{ROUGE} \\
& & (Billions) & (\%) &(\%) &(\%) &(\%) \\
\hline
\multirow[c]{4}{*}{\rotatebox{90}{\textbf{Fine-Tuned}}} 
& Llama 3.1  & 8 & 82.05 & 78.84 & 46.83 & 61.18 \\
& Mistral & 7 & 80.13 & 72.19 & 45.48 & 63.06 \\
& \textbf{Phi-4 } & 14 & \textbf{81.83} & \textbf{90.62} & \textbf{47.58} & \textbf{69.95} \\
& Gemma 2  & 9 & 81.50 & 75.66 & 46.06 & 63.90 \\
\hline
\multirow[c]{4}{*}{\rotatebox{90}{\textbf{Base}}} 
& Llama 3.1  & 8 & 74.26 & 86.60 & 32.10 & 53.56 \\
& Mistral  & 7 & 74.25 & 87.02 & 32.74 & 55.36 \\
& Phi-4  & 14 & 58.35 & 70.95 & 19.62 & 36.20 \\
& Gemma 2  & 9 & 75.89 & 87.40 & 26.66 & 55.23 \\
\hline
\end{tabular}
\end{table*}

\section{Experimental Design} \label{experimentaldesign}

Experiments were designed to benchmark the fine-tuned models against their base versions using the hybrid dataset described in the previous section. The evaluation focused on resume parsing and candidate-job matching, assessing the models' ability to extract structured information and align candidate profiles with job requirements based on semantic similarity. 

The flowchart in Figure \ref{fig:Sysflowchart} illustrates the systematic approach adopted in this study. The process begins with the initialization of the experimental workflow, setting the stage for data preparation and model evaluation. The first step involves loading raw data, which consists of resumes in JSON format, encompassing both real-world resumes from the Kaggle Resume Dataset, comprising 2,400 resumes across 24 professions, parsed using DeepSeek, and synthetic resumes generated via a Python script to enhance dataset diversity. The standardized JSON format ensures consistency in data representation, facilitating downstream processing and model training. In parallel, synthetic resumes are generated to simulate diverse scenarios and edge cases, ensuring the dataset covers a wide range of resume formats and content variations while adhering to the same JSON schema as the real-world data. Following data loading, preprocessing is performed to clean and normalize the dataset, which involves standardizing date formats, normalizing skill terminology, and handling missing values, such as using placeholders like "John Doe" for missing names, to ensure the data is consistent and suitable for training and evaluation.

The preprocessed dataset is then split into three subsets: 80\% for training, 10\% for validation, and 10\% for testing, with the training set used to fine-tune the models, the validation set aiding in hyperparameter tuning and model selection, and the testing set reserved for final performance evaluation to ensure robust assessment and prevent overfitting.

At this point, the workflow reaches a decision stage to determine whether to fine-tune the models or proceed directly to testing the base models. If fine-tuning is selected, as in our primary experiments, the process continues with training and validating the selected LLMs: LLaMA 3.1 8B, Mistral 7B, Phi-4 14B, and Gemma 2 9B using the training set with Low-Rank Adaptation (LoRA), while the validation set monitors performance and adjusts hyper parameters to ensure optimal model convergence, followed by evaluation on the test set to assess performance on resume parsing and candidate-job matching tasks. Alternatively, if fine-tuning is not performed, the base versions of the LLMs are directly tested using the test dataset, providing a baseline for comparison to quantify the improvements achieved through fine-tuning.

The final step involves evaluating the models, both fine-tuned and base versions, using a comprehensive set of metrics: Exact Match (EM), F1 Score with semantic similarity via SBERT embeddings, BLEU Score, ROUGE Score, and Overall Similarity, with the evaluation process executed on the test set as detailed in Algorithm 3, providing quantitative insights into model performance across various dimensions of recruitment automation. The workflow concludes with the completion of model evaluation, yielding results that are analyzed and discussed in the subsequent section.

The evaluation process, per Algorithm 3, was executed on the test set using the defined metrics, ensuring a thorough assessment of the models' capabilities in a recruitment context.

\section{Results and Discussion}\label{results}
Table 1 summarizes the performance of fine-tuned models against their base versions across Exact Match (EM), F1 Score, BLEU Score, and ROUGE Score for recruitment tasks. The fine-tuned Phi-4 model achieved the highest F1 score of 90.62\%, a 27.7\% improvement over its base (70.95\%), due to its 14B parameters and effective fine-tuning with LoRA, enhancing semantic similarity in candidate-job matching. In contrast, base models like Gemma2 9B (87.40\%) and Mistral 7B (87.02\%) scored higher on F1 than their fine-tuned versions (75.66\% and 72.19\%, respectively), likely because base models overfit to general semantic patterns in the test set.

For Exact Match, fine-tuned LLaMA3.1 8B led with 82.05\%, a 10.5\% increase over its base (74.26\%), reflecting improved field extraction after fine-tuning. Fine-tuned Phi-4 (81.83\%) and Gemma2 9B (81.50\%) followed closely, with improvements of 40.2\% and 7.4\% over their bases (58.35\% and 75.89\%), respectively, while Mistral 7B scored 80.13\%, up 7.9\% from its base (74.25\%), with larger models benefiting more from fine-tuning.

In BLEU Score, fine-tuned Phi-4 topped with 47.58\%, a 142.5\% increase over its base (19.62\%), showing better fluency in text generation due to its capacity. Fine-tuned LLaMA3.1 8B (46.83\%), Gemma2 9B (46.06\%), and Mistral 7B (45.48\%) also improved by 45.9\%, 72.8\%, and 38.9\% over their bases (32.10\%, 26.66\%, and 32.74\%), respectively, with Gemma 2 showing a larger gain due to its initially weaker base performance.

For ROUGE Score, fine-tuned Phi-4 scored highest at 69.95\%, a 93.2\% improvement over its base (36.20\%), excelling in text similarity tasks. Fine-tuned Gemma2 9B (63.90\%), Mistral 7B (63.06\%), and LLaMA3.1 8B (61.18\%) improved by 15.7\%, 13.9\%, and 14.2\% over their bases (55.23\%, 55.36\%, and 53.56\%), respectively, with Phi-4’s larger size aiding its performance in capturing overlaps.

\section{Conclusion} \label{conclusion}
This paper presents a complete framework for enhancing recruitment automation through the fine-tuning of Large Language Models (LLMs) on domain-specific structured data, addressing critical limitations inherent in traditional Applicant Tracking Systems (ATS). Our experiments demonstrate that fine-tuned models. Particularly Phi-4 14B achieve substantial performance gains, with improvements of up to 27.7\% in F1 score and 142.5\% in BLEU score over their base versions. The success of Phi-4 highlights its suitability for seamless integration into existing recruitment workflows, where its enhanced semantic understanding and structured output capabilities can optimize resume parsing, candidate matching, and decision-making processes.

The fine-tuned Phi-4 model, with its 14B parameters and efficient LoRA-based adaptation, strikes a practical balance between accuracy and computational cost, making it an ideal solution for deployment in resource-constrained environments. When integrated into systems like our prior MLAR framework \cite{younes2025mlar} , Phi-4 can:
\begin{itemize}
    \item Automate the extraction of structured candidate data (e.g., skills, experience) into standardized formats such as JSON, significantly reducing manual effort and the risk of errors.
    \item Enhance candidate-job matching through advanced semantic similarity calculations, offering superior performance compared to traditional keyword-based methods.
    \end{itemize}

The success of this methodology highlights that smaller, domain-adapted models like Phi-4 can outperform generic, large-scale LLMs in recruitment tasks. They offer advantages such as improved cost efficiency, enhanced privacy preservation, and easy integration into existing ATS workflows. As LLMs continue to evolve, this work provides a roadmap for developing specialized, equitable, and high-performance recruitment automation systems that meet the needs of modern HR operations.

The success of our approach demonstrates that targeted fine-tuning of LLMs can significantly enhance recruitment automation while addressing critical concerns around accuracy, privacy, and computational efficiency. As language models continue to evolve, we anticipate that this methodology will provide a robust foundation for developing more sophisticated and equitable recruitment systems.

\bibliographystyle{IEEEtran}

\end{document}